\documentclass[conference,hidelinks]{IEEEtran}
\IEEEoverridecommandlockouts

\usepackage{amsmath}
\usepackage{amssymb}
\usepackage{amsthm}
\usepackage{booktabs}
\usepackage{cite}
\usepackage{color}
\usepackage{enumitem}
\usepackage{graphicx}
\usepackage{hyperref}
\usepackage{pifont}
\newcommand{\cmark}{\ding{51}}%
\newcommand{\xmark}{\ding{55}}%

\newcommand\fnurl[2]{\href{#2}{#1}\footnote{\url{#2}}}

\IEEEoverridecommandlockouts
\IEEEpubid{\begin{minipage}[t]{1.1\textwidth}\ \\[10pt]
        \centering{\fontsize{6.5}{7} \selectfont \copyright 2019 IEEE. Personal use of this material is permitted. Permission from IEEE must be obtained for all other uses, in any current or future media, including reprinting/republishing this material for advertising or promotional purposes, creating new collective works, for resale or redistribution to servers or lists, or reuse of any copyrighted component of this work in other works}
\end{minipage}}

\begin{document}

\title{Do we still need fuzzy classifiers for Small Data in the Era of Big Data?
\thanks{This work has been supported by the Spanish Ministry of Economy and Competitiveness under the project TIN2016-77356-P (MINECO, AEI/FEDER, UE) and by the Public University of Navarra under the project PJUPNA13.}
}

\author{\IEEEauthorblockN{Mikel~Elkano}
\IEEEauthorblockA{Institute of Smart Cities \\
Public University of Navarre\\
Pamplona, Spain \\
mikel.elkano@unavarra.es}
\and
\IEEEauthorblockN{Humberto~Bustince}
\IEEEauthorblockA{Institute of Smart Cities \\
Public University of Navarre\\
Pamplona, Spain \\
bustince@unavarra.es}
\and
\IEEEauthorblockN{Mikel~Galar}
\IEEEauthorblockA{Institute of Smart Cities \\
Public University of Navarre\\
Pamplona, Spain \\
mikel.galar@unavarra.es}
}

\maketitle

\begin{abstract}
The Era of Big Data has forced researchers to explore new distributed solutions for building fuzzy classifiers, which often introduce approximation errors or make strong assumptions to reduce computational and memory requirements. As a result, Big Data classifiers might be expected to be inferior to those designed for standard classification tasks (Small Data) in terms of accuracy and model complexity. To our knowledge, however, there is no empirical evidence to confirm such a conjecture yet. Here, we investigate the extent to which state-of-the-art fuzzy classifiers for Big Data sacrifice performance in favor of scalability. To this end, we carry out an empirical study that compares these classifiers with some of the best performing algorithms for Small Data. Assuming the latter were generally designed for maximizing performance without considering scalability issues, the results of this study provide some intuition around the tradeoff between performance and scalability achieved by current Big Data solutions. Our findings show that, although slightly inferior, Big Data classifiers are gradually catching up with state-of-the-art classifiers for Small data, suggesting that a unified learning algorithm for Big and Small Data might be possible.
\end{abstract}

\section{Introduction}
Fuzzy logic has allowed machine learning algorithms to improve the tradeoff between classification performance and model interpretability. Fuzzy classifiers are not only able to explain their predictions with human-readable linguistic labels~\cite{Ishibuchi2004}, but also to improve the classification performance of non-fuzzy methods\cite{Huhn2009,Segatori2018}. Among the features that make them stand out from other types of classifiers is their ability to deal with uncertainty and create soft decision boundaries. However, the construction of interpretable models usually involves computationally intensive learning algorithms that require long runtimes. In the Era of Big Data, researchers have been forced to design distributed algorithms able to run on computing clusters that meet the minimum computational and memory requirements, often using open-source frameworks such as \fnurl{Apache Hadoop}{http://hadoop.apache.org} and \fnurl{Apache Spark}{https://spark.apache.org} \cite{Elkano2017, Elkano2019, Fernandez2017, Lopez2015, Segatori2018}.

Although existing Big Data algorithms have shown promising results, the extent to which they sacrifice performance (both accuracy and model complexity) in favor of scalability is not clear. In general, scalable solutions usually rely on locally optimal subsolutions which might introduce approximation errors into the learning process or make strong assumptions about the training data which might not hold. These strategies help reduce computational and memory requirements but might decrease performance. This means that Big Data algorithms might be inferior to the state-of-the art classifiers designed for Small Data in terms of accuracy and model complexity\footnote{For the sake of simplicity, we will use the term \textit{Small Data} to refer to standard classification tasks}. Since Small Data classifiers are generally designed for maximizing performance without considering scalability issues, comparing Big vs. Small Data algorithms on Small Data provides some intuition around the tradeoff between performance and scalability achieved by Big Data solutions.

In this work, we carry out an empirical study consisting of 18 Small Data classification tasks available at UCI~\cite{Lichman2013}, KEEL~\cite{AlcalaFdez2011b}, and \fnurl{OpenML}{https://www.openml.org/search?type=data} repositories. Among the different types of fuzzy classifiers, we have focused on Fuzzy Rule-Based Classification Systems (FRBCSs)~\cite{Ishibuchi2004} and Fuzzy Decision Trees (FDTs)~\cite{Yuan1995}, since they usually provide a good accuracy-interpretability tradeoff. We consider only open-source implementations of Big and Small Data algorithms. We measure performance in terms of accuracy rate and model complexity considering the average number of rules (leaves), antecedents (depth of the tree), and fuzzy sets per variable.

This paper is organized as follows. Section \ref{sec:motivation} explains the motivation of this work along with the basics of FRBCSs and FDTs. In Section \ref{sec:study}, we describe the data and methods used in the experimental study and present the results. Finally, Section \ref{sec:discussion} concludes this paper.

\section{Motivation}\label{sec:motivation}
In this work, we try to answer the following questions:
\begin{itemize}
\item Can fuzzy classifiers designed for Big Data achieve state-of-the-art performance on standard (Small Data) classification tasks?
\item Have Big Data solutions unseated Small Data ones, or do we still need to sacrifice classification performance in favor of scalability?
\end{itemize}

Providing answers to these questions is important to understand whether current Big Data solutions are introducing significant approximation errors and/or making too strong assumptions about training data to save computational and memory resources. Analyzing these two aspects would help indicate whether we are on the right track towards achieving scalable learning algorithms.

\subsection{Fuzzy classifiers}
In particular, we focus on one of the main benefits that fuzzy classifiers usually bring: a good tradeoff between classification performance and interpretability. Among the different types of classifiers, we consider the Fuzzy Rule-Based Classification Systems (FRBCSs)~\cite{Ishibuchi2004}, which build models based on human-readable IF-THEN rules composed of linguistic labels. The two main components of FRBCSs are the following:
\begin{enumerate}
\item Knowledge base (KB): it is composed of both the rule base (RB) and the database (DB), where the rules and membership functions used to model the linguistic labels are stored, respectively.
\item Fuzzy Reasoning Method (FRM): this is the mechanism used to classify examples with the information stored in the KB.
\end{enumerate}

In addition to FRBCSs, we consider Fuzzy Decision Trees (FDTs)~\cite{Yuan1995}, which make use of fuzzy logic to better deal with uncertainty and create soft decision boundaries that improve classification performance. FDTs use fuzzy partitions to characterize continuous attributes instead of considering a discrete set of intervals, and thus multiple branches can be activated simultaneously. For the sake of readability, we will use the term \textit{fuzzy classifier} to refer only to FRBCSs and FDTs.

\subsection{Fuzzy classifiers for Big Data}
In general, the distributed algorithms proposed for Big Data so far consist either in applying a divide-and-conquer strategy based on multiple local optimization sub-problems~\cite{Lopez2015,Fernandez2017} or in performing a global distributed learning process~\cite{Elkano2017,Segatori2018}. In the former case, the final classifier is built by aggregating several independent models obtained in different subsets of data. In this case, the learning process becomes strongly dependent on the distribution of subsets and might miss important information available only when training data is treated as a whole. Regarding global distributed algorithms, the difficulty of parallelizing the learning stage across several computing units is the main drawback.

In this work, we aim to assess the extent to which the best-performing fuzzy classifiers for Big Data are sacrificing performance (both accuracy and model complexity) in favor of scalability.

\section{Experimental study}\label{sec:study}
We carried out an empirical study to test a number of state-of-the-art fuzzy classifiers designed for Big Data on Small Data and analyze how they perform in comparison with some of the most accurate Small Data classifiers.

\subsection{Data and methods}\label{ssec:framework}
We analyzed the performance of each algorithm on 18 Small Data classification tasks available at UCI~\cite{Lichman2013}, KEEL~\cite{AlcalaFdez2011b}, and \fnurl{OpenML}{https://www.openml.org/search?type=data} repositories. Table \ref{tab:datasets} shows the description of these datasets based on the number of examples, classes, and features (R = real, I = integer, C = categorical). The performance of all methods was assessed with a \textit{5-fold stratified cross-validation scheme}, where each dataset is randomly split into five equal-sized partitions of data and the model is trained with a combination of four of them (80\%) and tested with the remaining partition. Therefore, the result of each dataset was computed as the average of the five partitions.

\begin{table}[htbp]
 	\centering
 	\caption{description of the datasets.}
 	\renewcommand\tabcolsep{3pt}
	\renewcommand{\arraystretch}{1.2}
    \begin{tabular}{@{}llrrrrrr@{}}
        \toprule
        \multicolumn{1}{c}{ID} & \multicolumn{1}{c}{Name} & \multicolumn{1}{c}{\#Examples} & \multicolumn{1}{c}{\#Classes} & \multicolumn{4}{c}{\#Features}\\
        & & & & \multicolumn{1}{c}{Total} & \multicolumn{1}{c}{R} & \multicolumn{1}{c}{I} & \multicolumn{1}{c}{C}\\
        \midrule
        ADULT & Adult & 45,222 & 2 & 14 & 6 & 0 & 8 \\
        CLEVE & Cleveland & 297 & 5 & 13 & 13 & 0 & 0 \\
        CONTR & Contraceptive & 1,473 & 3 & 9 & 6 & 0 & 3 \\
        CRX & CRX & 653 & 2 & 15 & 3 & 3 & 9 \\
        IONOS & Ionosphere & 351 & 2 & 33 & 32 & 1 & 0 \\
        MAGIC & Magic & 19,020 & 2 & 10 & 10 & 0 & 0 \\
        MAMMO & Mammographic & 830 & 2 & 5 & 0 & 5 & 0 \\
        NEWTH & Newthyroid & 215 & 3 & 5 & 5 & 0 & 0 \\
        PAGEB & Pageblocks & 548 & 5 & 10 & 10 & 0 & 0 \\
        PENBA & Penbased & 1,100 & 10 & 16 & 16 & 0 & 0 \\
        PHONE & Phoneme & 5,404 & 2 & 5 & 5 & 0 & 0 \\
        RING & Ring & 7,400 & 2 & 20 & 20 & 0 & 0 \\
        SHUTT & Shuttle & 2,175 & 7 & 9 & 0 & 9 & 0 \\
        THYRO & Thyroid & 720 & 3 & 21 & 6 & 15 & 0 \\
        TITAN & Titanic & 2,201 & 2 & 3 & 3 & 0 & 0 \\
        VEHIC & Vehicle & 846 & 4 & 18 & 18 & 0 & 0 \\
        WDBC & WDBC & 569 & 2 & 30 & 30 & 0 & 0 \\
        WISCO & Wisconsin & 683 & 2 & 9 & 0 & 9 & 0 \\
        \bottomrule
    \end{tabular}%
 	\label{tab:datasets}%
\end{table}%

We considered all the open-source fuzzy classifiers available for Big Data so far\footnote{Chi-FRBCS-BigData~\cite{Lopez2015} was omitted because CHI-BD showed better performance in~\cite{Elkano2017}} (CHI-BD~\cite{Elkano2017}, Chi-Spark-RS~\cite{Fernandez2017}, CFM-BD~\cite{Elkano2019}, and FBDT/FMDT~\cite{Segatori2018}) and two of the best-performing fuzzy classifiers for Small Data (FARC-HD~\cite{AlcalaFdez2011a} and FURIA~\cite{Huhn2009}). Although the models and learning algorithms used by FRBCSs and FDTs are different, the leaves of FDTs can be converted into a set of IF-THEN rules, allowing us to compare the accuracy-interpretability tradeoff of both types of classifiers. Next, we briefly describe each of these algorithms to better understand their behavior throughout the experiments:

\begin{itemize}

\item \textit{CHI-BD}~\cite{Elkano2017} (for Big Data): this method recovers the original Chi et al. algorithm in Big Data without any approximation error. Contrary to previous approaches~\cite{Lopez2015}, CHI-BD generates a single rule base using the whole training set instead of fusing independent rule bases.

\item \textit{Chi-Spark-RS}~\cite{Fernandez2017} (for Big Data): this algorithm optimizes the rule base obtained with Chi-FRBCS-BigData~\cite{Lopez2015} by introducing an evolutionary optimization stage. Chi-FRBCS-BigData learns multiple locally-optimal rule bases by applying independent Chi et al. classifiers on disjoint partitions of the training set. The final rule base is generated by aggregating all the local rule bases, which implies there often exists some approximation error that tends to be higher as the number of partitions increases.

\item \textit{CFM-BD}~\cite{Elkano2019} (for Big Data): this fuzzy rule induction algorithm was designed for building compact models that maximize the accuracy-interpretability tradeoff. The learning process consists of three sequential stages:
\begin{enumerate}
\item Pre-processing and partitioning: the shape and position of the fuzzy sets are adjusted to the real distribution of training data.
\item Rule induction process: rules are constructed with an algorithm inspired by CHI-BD~\cite{Elkano2017} and Apriori~\cite{Agrawal1994}. The induction process consists in finding the most frequent itemsets (sets of linguistic labels) and selecting the rules with the greatest discrimination capability.
\item Evolutionary rule selection: a distributed version of the CHC evolutionary algorithm~\cite{Eshelman1991} is used for selecting the most accurate rules.
\end{enumerate}

\item \textit{FBDT/FMDT}~\cite{Segatori2018} (for Big Data): Segatori et al. proposed a distributed fuzzy decision tree (FDT) that extends the implementation of decision trees in \fnurl{Spark MLlib}{http://spark.apache.org/mllib}. This method comprises two stages:
\begin{enumerate}
\item Fuzzy partitioning: a strong triangular fuzzy partition is built for each continuous attribute based on fuzzy entropy, which is then used to construct the tree.
\item FDT learning: the tree is constructed with one of the two versions of FDT proposed by the authors, which differ in the splitting strategy: the binary (or two-way) FDT (FBDT) and the multi-way FDT (FMDT). The former recursively partitions the attribute space into two subspaces (child nodes), while the latter might generate more than two subspaces.
\end{enumerate}

\item \textit{FARC-HD}~\cite{AlcalaFdez2011a} (for Small Data): this fuzzy association rule-based classifier applies a rule induction process comprising three stages:
\begin{enumerate}
\item Fuzzy association rule extraction: a search tree is constructed for each class to extract frequent itemsets with the Apriori algorithm. Once the frequent itemsets are obtained, the fuzzy rules are extracted.
\item Candidate rule pre-screening: the most interesting fuzzy rules are selected with a pattern weighting scheme based on the coverage of the fuzzy rules.
\item Genetic rule selection and lateral tuning: an evolutionary algorithm tunes the lateral position of the membership functions and selects the most accurate rules.
\end{enumerate} 

\item \textit{FURIA}~\cite{Huhn2009} (for Small Data): this algorithm modifies and extends the RIPPER rule induction algorithm~\cite{Cohen1995}. To extract fuzzy rules, FURIA applies the following procedure:
\begin{enumerate}
\item Learn a rule set for each class using the RIPPER algorithm.
\item Fuzzify the rules generated by RIPPER: the interval representing each antecedent is replaced by a trapezoidal membership function which is optimized by means of a greedy algorithm.
\end{enumerate}

\end{itemize}

Although all these classifiers build models composed of IF-THEN rules, the rule structure may vary among them. Table \ref{tab:rules} shows these differences considering the following aspects:
\begin{itemize}
\item Rule length: the majority of methods build rules of variable length. Short rules generally provide greater generalization power and make the model more interpretable.
\item Duplicated antecedents: sometimes attributes appear more than once in the same antecedent part (associated with different fuzzy sets), which might affect interpretability.
\item Trainable fuzzy sets: some fuzzy partitioning methods construct the fuzzy sets based on the training set, adjusting their position and/or shape, and even the number of fuzzy sets used for each variable.
\item Shared fuzzy sets: in some cases each rule might use its own fuzzy sets, which can significantly increase the number of generated fuzzy sets and affect interpretability.
\end{itemize}
\begin{table}[!htbp]
	\centering
	\caption{Differences in the rule structure used by each algorithm.}
	\begin{tabular}{@{}lcclc@{}}
		\toprule
		& \multicolumn{1}{c}{Rule} & \multicolumn{1}{c}{Dupli.} & & \multicolumn{1}{c}{Shared} \\
		Algorithm & \multicolumn{1}{c}{Length} & \multicolumn{1}{c}{Ants.} & \multicolumn{1}{l}{Trainable FS} & \multicolumn{1}{c}{FS}\\
		\midrule
		CHI-BD & \#features & \xmark & No (fixed) & \cmark \\
		Chi-Spark-RS & \#features & \xmark & No (fixed) & \cmark \\
		CFM-BD & Variable & \xmark & Position and shape & \cmark \\
		FBDT & Variable & \cmark & \#FS, position, and shape & \cmark \\
		FMDT & Variable & \xmark & \#FS, position, and shape & \cmark \\
		FARC-HD & Variable & \xmark & Position & \cmark \\
		FURIA & Variable & \xmark & \#FS, position, and shape & \xmark \\
		\bottomrule
		\multicolumn{5}{l}{FS stands for ``fuzzy sets''.}\\
	\end{tabular}
	\label{tab:rules}
\end{table}

Regarding the parameters used for each method (Table \ref{tab:parameters}), we set their values based on the recommendations from the authors. Although some of them have a cost-sensitive mode for imbalanced datasets, we disabled this mode for all methods to perform fair comparisons based on the accuracy rate when assessing classification performance. For Big Data algorithms, we used only one executor/partition for all datasets.

\begin{table}[!htbp]
	\centering
	\caption{parameters used for each method.}
	\begin{tabular}{@{}ll@{}}
		\toprule
		Algorithm & Parameters \\
		\midrule
		& \#Fuzzy sets per variable = 3\\
		& Inference = winning rule \\
		CHI-BD & Rule weight = certainty factor\\
		& \#Rule subsets = 4\\
		& Min. \#occurrences for frequent subsets = 10\\
		& Max. \#rules per reducer = 400,000\\
		\midrule
		& \#Fuzzy sets per variable = 3\\
		Chi-Spark-RS & Inference = winning rule \\
		& Rule weight = certainty factor\\
		& \#Individuals = 50; \#Evaluations = 1,000; $\alpha$ = 0.7\\
		\midrule
		& \#Fuzzy sets per variable = 5\\
		& Inference = winning rule \\
		& Rule weight = certainty factor\\
		CFM-BD & $maxLen$ = 3; $prop$ = (0.2, 0.3, 0.5) \\
		& $minConf_{crisp}$ = 0.7; $minConf_{fuzzy}$ = 0.6 \\
		& $\gamma$ = 4; $\delta$ = 0.15; $\Gamma$ = 0.35; $\varphi$ = 0.01 \\
		& \#Individuals = 50; \#Evaluations = 10,000 \\
		& $maxRestarts$ = 3; $D$ = $NR_{initial}$ / 4 \\
		\midrule
		& Impurity = entropy; T-norm = product \\
		FBDT & $maxBins$ = 32; $maxDepth$ ($\beta$) = 5 \\
		& $\gamma$ = 0.1\%; $\phi$ = 1; $\lambda$ = 1 \\
		\midrule
		& Impurity = entropy; T-norm = product \\
		FMDT & $maxBins$ = 32; $maxDepth$ ($\beta$) = 5 \\
		& $\gamma$ = 0.1\%; $\phi$ = 0.02 $\cdot$ $N$; $\lambda$ = 10$^{-4}\cdot N$ \\
		\midrule
		 & \#Fuzzy sets per variable = 5\\
		 & Inference = additive combination\\
		 & Rule weight = certainty factor\\
		 FARC-HD & Min. support = 0.05; Min. confidence = 0.8\\
		 & Max. depth = 3; $k$ = 2\\
		 & \#Individuals = 50; \#Evaluations: 20,000; $\alpha$ = 0.02\\
		 & Bits per gen = 30\\
		\midrule
		 FURIA & \#Optimizations = 2\\
		 & \#Folds = 3 \\
		\bottomrule
	\end{tabular}
	\label{tab:parameters}
\end{table}

\subsection{Analysis of the results}\label{ssec:analysis}
Tables \ref{tab:accuracy-small-data}, \ref{tab:complexity-big-small-data} and \ref{tab:complexity-small-data} show the accuracy rates and model complexities of each algorithm, respectively. To evaluate model complexity, we considered the average number of rules \#rules (leaves), the average number of antecedents \#ants. (depth of the tree), and the average number of fuzzy sets per variable \#FS.

\vspace{5pt}
\subsubsection{Performance of Big Data algorithms on Small Data}\label{sssec:analysis-big-data}
\indent According to these results and the experimental study presented in~\cite{Elkano2019}, Big Data classifiers followed similar trends on Big and Small Data. While FMDT was the most accurate method, the resulting trees were significantly more complex than the models built by CFM-BD and FBDT. In terms of model complexity, CFM-BD and FBDT provided the most compact models while achieving competitive, though slightly inferior, classification performance with respect to FMDT. Both methods (CFM-BD and FBDT) generated a similar number of rules but used a different number of antecedents and fuzzy sets. Regarding the antecedents, rules built by CFM-BD are more general (shorter) than those extracted by FBDT and they do not contain duplicated antecedents (contrary to the rules of FBDT), which improves interpretability. As for fuzzy sets, both FBDT and FMDT applies a partitioning method that adjusts the number of fuzzy sets for each variable based on the training set, which allowed them to use less fuzzy sets than CFM-BD on Small Data. However, we must remark that the experiments in~\cite{Elkano2019} showed that CFM-BD was able to deal with Big Data problems using less fuzzy sets than FBDT and FMDT, which suggests that CFM-BD might achieve competitive classification performance on Small Data with less fuzzy sets. 

Exceptions to these trends were PENBA and RING datasets, where CFM-BD built significantly more rules than FBDT. In the case of PENBA, the number of classes (10) caused CFM-BD to create more rules to distinguish all the classes, since this method was designed for maximizing classification performance for all classes instead of the accuracy rate. As for RING, FBDT built less rules than CFM-BD but sacrificed classification performance reducing the accuracy rate by \%11 with respect to CFM-BD. Apparently, CFM-BD might be the algorithm offering the best accuracy-complexity tradeoff on Big and Small Data among the Big Data algorithms considered in this study.

Regarding Chi-based methods (CHI-BD and {Chi-Spark-RS}), they were clearly outperformed by CFM-BD and FBDT/FMDT in terms of accuracy and complexity, and hence we omitted these two methods in the subsequent analysis.

\begin{table*}[!htbp]
    \centering
    \caption{classification accuracy (\%) comparison.}
    \renewcommand\tabcolsep{8pt}
    \renewcommand{\arraystretch}{1.3}
    \begin{tabular}{@{}lccccccc@{}}
        \toprule
        Dataset & CHI-BD & Chi-Spark-RS$^*$ & CFM-BD & FBDT & FMDT & FARC-HD & FURIA \\
		\midrule
        ADULT & 70.31 & 61.97 & 82.86 & 84.26 & \textbf{84.47} & 83.55 & 83.28 \\
        CLEVE & 54.20 & - & 54.20 & 54.89 & 54.21 & \textbf{57.80} & 56.57 \\
        CONTR & 47.18 & - & 49.42 & 53.09 & 53.22 & 53.63 & \textbf{54.17} \\
        CRX & 69.22 & 65.70 & 85.31 & 84.39 & 82.24 & \textbf{86.53} & 86.37 \\
        IONOS & 67.84 & 65.26 & 88.90 & 84.93 & 88.03 & \textbf{90.32} & 88.91 \\
        MAGIC & 77.08 & 78.79 & 83.87 & 81.24 & \textbf{84.84} & 84.51 & 84.83 \\
        MAMMO & 81.38 & 81.39 & 83.57 & 81.55 & 81.04 & \textbf{84.19} & 83.57 \\
        NEWTH & 85.12 & - & 91.16 & 91.63 & \textbf{95.35} & 95.04 & 94.88 \\
        PAGEB & 91.42 & - & 94.16 & 95.43 & \textbf{95.98} & 94.18 & 95.25 \\
        PENBA & \textbf{94.00} & - & 84.82 & 84.73 & 91.18 & 93.05 & 92.45 \\
        PHONE & 71.95 & 77.13 & 79.79 & 78.76 & 80.29 & 81.37 & \textbf{85.90} \\
        RING & 55.27 & 82.18 & 90.59 & 79.61 & 86.78 & 93.62 & \textbf{94.01} \\
        SHUTT & 80.23 & - & 99.54 & 99.08 & 99.17 & 95.50 & \textbf{99.68} \\
        THYRO & 87.92 & - & 91.39 & 95.97 & 95.97 & 93.52 & \textbf{98.47} \\
        TITAN & 67.70 & 76.92 & 77.47 & 67.70 & 67.70 & \textbf{78.87} & 78.51 \\
        VEHIC & 60.76 & - & 65.13 & 57.56 & 64.89 & 69.90 & \textbf{70.21} \\
        WDBC & 93.15 & 92.62 & 90.68 & 94.55 & 95.78 & \textbf{96.49} & 95.78 \\
        WISCO & 90.78 & 87.56 & 96.04 & 96.48 & \textbf{97.22} & 96.36 & 96.19 \\
        \midrule
        AVG & 74.75 & 76.95 & 82.72 & 81.44 & 83.24 & 84.91 & \textbf{85.50} \\
        \bottomrule
        \multicolumn{8}{l}{$^*$The source code of Chi-Spark-RS available at GitHub had no multi-class support.}\\
    \end{tabular}%
    \label{tab:accuracy-small-data}%
\end{table*}%
\begin{table*}[!htbp]
    \centering
    \caption{model complexity of the algorithms designed for big data.}
    \renewcommand\tabcolsep{6pt}
    \renewcommand{\arraystretch}{1.3}
    \begin{tabular}{lrrrrrrrrrrrrrrr}
        \toprule
        Dataset & \multicolumn{3}{c}{CHI-BD} & \multicolumn{3}{c}{Chi-Spark-RS$^*$} & \multicolumn{3}{c}{CFM-BD} & \multicolumn{3}{c}{FBDT} & \multicolumn{3}{c}{FMDT}\\
        & \#rules & \#ants. & \#FS & \#rules & \#ants. & \#FS & \#rules & \#ants. & \#FS & \#rules & \#ants. & \#FS & \#rules & \#ants. & \#FS \\
        \cmidrule(r){1-1}\cmidrule(r){2-4}\cmidrule(r){5-7}\cmidrule(r){8-10}\cmidrule(r){11-13}\cmidrule(r){14-16}
        ADULT & 13,225.00 & 14.00 & 3.00 & 6,606.80 & 14.00 & 3.00 & 15.20 & 1.58 & 5.00 & 29.40 & 4.92 & 7.77 & 3,295.40 & 4.77 & 7.77 \\
        CLEVE & 204.00 & 13.00 & 3.00 & \multicolumn{1}{l}{-} & \multicolumn{1}{l}{-} & \multicolumn{1}{l}{-} & 14.20 & 2.80 & 5.00 & 8.40 & 3.05 & 0.46 & 10.80 & 2.00 & 0.46 \\
        CONTR & 238.00 & 9.00 & 3.00 & \multicolumn{1}{l}{-} & \multicolumn{1}{l}{-} & \multicolumn{1}{l}{-} & 16.60 & 2.84 & 5.00 & 28.80 & 4.90 & 1.80 & 95.60 & 4.67 & 1.80 \\
        CRX  & 396.40 & 15.00 & 3.00 & 206.40 & 15.00 & 3.00 & 21.40 & 2.19 & 5.00 & 19.60 & 4.64 & 1.37 & 122.40 & 4.00 & 1.37 \\
        IONOS & 204.20 & 33.00 & 3.00 & 112.60 & 33.00 & 3.00 & 27.40 & 2.05 & 5.00 & 20.60 & 4.61 & 3.14 & 177.00 & 2.73 & 3.14 \\
        MAGIC & 298.60 & 10.00 & 3.00 & 116.00 & 10.00 & 3.00 & 16.20 & 1.93 & 5.00 & 23.80 & 4.74 & 6.08 & 2,923.60 & 3.64 & 6.08 \\
        MAMMO & 42.00 & 5.00 & 3.00 & 4.80 & 5.00 & 3.00 & 6.60 & 1.91 & 5.00 & 24.80 & 4.71 & 2.40 & 59.40 & 3.93 & 2.40 \\
        NEWTH & 17.80 & 5.00 & 3.00 & \multicolumn{1}{l}{-} & \multicolumn{1}{l}{-} & \multicolumn{1}{l}{-} & 11.80 & 1.93 & 5.00 & 19.60 & 4.53 & 3.64 & 94.40 & 3.12 & 3.64 \\
        PAGEB & 19.00 & 10.00 & 3.00 & \multicolumn{1}{l}{-} & \multicolumn{1}{l}{-} & \multicolumn{1}{l}{-} & 4.00 & 2.47 & 5.00 & 27.60 & 4.87 & 2.56 & 115.00 & 2.87 & 2.56 \\
        PENBA & 615.40 & 16.00 & 3.00 & \multicolumn{1}{l}{-} & \multicolumn{1}{l}{-} & \multicolumn{1}{l}{-} & 180.40 & 2.85 & 5.00 & 31.80 & 4.99 & 3.45 & 355.20 & 3.76 & 3.45 \\
        PHONE & 49.40 & 5.00 & 3.00 & 9.20 & 5.00 & 3.00 & 8.80 & 2.11 & 5.00 & 22.00 & 4.69 & 5.56 & 1,302.80 & 3.23 & 5.56 \\
        RING & 515.40 & 20.00 & 3.00 & 425.40 & 20.00 & 3.00 & 51.80 & 1.54 & 5.00 & 18.40 & 4.56 & 6.17 & 2,847.60 & 2.63 & 6.17 \\
        SHUTT & 6.60 & 9.00 & 3.00 & \multicolumn{1}{l}{-} & \multicolumn{1}{l}{-} & \multicolumn{1}{l}{-} & 6.00 & 2.13 & 5.00 & 23.60 & 4.69 & 4.51 & 373.40 & 2.49 & 4.51 \\
        THYRO & 123.20 & 21.00 & 3.00 & \multicolumn{1}{l}{-} & \multicolumn{1}{l}{-} & \multicolumn{1}{l}{-} & 4.40 & 1.90 & 5.00 & 8.80 & 3.53 & 0.48 & 22.40 & 2.35 & 0.48 \\
        TITAN & 0.00 & 3.00 & 3.00 & 1.00 & 3.00 & 3.00 & 1.00 & 1.20 & 5.00 & 3.00 & 1.67 & 1.00 & 3.00 & 1.00 & 1.00 \\
        VEHIC & 213.00 & 18.00 & 3.00 & \multicolumn{1}{l}{-} & \multicolumn{1}{l}{-} & \multicolumn{1}{l}{-} & 31.00 & 2.21 & 5.00 & 23.80 & 4.71 & 2.97 & 320.40 & 3.59 & 2.97 \\
        WDBC & 380.20 & 30.00 & 3.00 & 124.20 & 30.00 & 3.00 & 19.40 & 1.11 & 5.00 & 25.60 & 4.81 & 2.76 & 280.60 & 3.20 & 2.76 \\
        WISCO & 202.80 & 9.00 & 3.00 & 82.60 & 9.00 & 3.00 & 7.80 & 1.78 & 5.00 & 23.40 & 4.82 & 3.24 & 181.80 & 4.67 & 3.24 \\
        \cmidrule(r){1-1}\cmidrule(r){2-4}\cmidrule(r){5-7}\cmidrule(r){8-10}\cmidrule(r){11-13}\cmidrule(r){14-16}
        AVG  & 930.61 & 13.61 & 3.00 & 768.90 & 14.40 & 3.00 & 24.67 & 2.03 & 5.00 & 21.28 & 4.41 & 3.30 & 698.93 & 3.26 & 3.30 \\
        \bottomrule
        \multicolumn{16}{l}{$^*$The source code of Chi-Spark-RS available at GitHub had no multi-class support.}\\
    \end{tabular}%
    \label{tab:complexity-big-small-data}%
\end{table*}%
\begin{table}[!htbp]
    \centering
    \caption{model complexity of the algorithms designed for small data.}
    \renewcommand\tabcolsep{7.5pt}
    \renewcommand{\arraystretch}{1.3}
    \begin{tabular}{lrrrrrr}
        \toprule
        Dataset & \multicolumn{3}{c}{FARC-HD} & \multicolumn{3}{c}{FURIA$^*$}\\
        & \#rules & \#ants. & \#FS & \#rules & \#ants. & \#FS\\
        \cmidrule(r){1-1}\cmidrule(r){2-4}\cmidrule(r){5-7}
        ADULT & 86.80 & 2.52 & 5.00 & 8.80 & 2.59 & - \\
        CLEVE & 62.27 & 2.90 & 5.00 & 7.20 & 2.89 & - \\
        CONTR & 64.60 & 2.68 & 5.00 & 8.40 & 2.70 & - \\
        CRX  & 26.60 & 2.58 & 5.00 & 7.40 & 2.52 & - \\
        IONOS & 16.20 & 2.03 & 5.00 & 11.00 & 2.40 & - \\
        MAGIC & 44.40 & 2.45 & 5.00 & 28.00 & 3.12 & - \\
        MAMMO & 20.00 & 1.85 & 5.00 & 3.40 & 1.57 & - \\
        NEWTH & 10.00 & 1.63 & 5.00 & 7.00 & 2.24 & - \\
        PAGEB & 13.27 & 2.43 & 5.00 & 11.00 & 2.34 & - \\
        PENBA & 78.40 & 2.80 & 5.00 & 39.80 & 3.66 & - \\
        PHONE & 18.00 & 2.18 & 5.00 & 30.20 & 3.93 & - \\
        RING & 25.40 & 1.92 & 5.00 & 83.60 & 4.89 & - \\
        SHUTT & 7.07 & 1.87 & 5.00 & 6.80 & 1.86 & - \\
        THYRO & 4.53 & 2.13 & 5.00 & 5.80 & 1.98 & - \\
        TITAN & 4.40 & 1.14 & 5.00 & 5.40 & 2.17 & - \\
        VEHIC & 44.87 & 2.66 & 5.00 & 21.60 & 3.29 & - \\
        WDBC & 11.60 & 1.62 & 5.00 & 9.60 & 2.54 & - \\
        WISCO & 13.60 & 1.22 & 5.00 & 12.80 & 2.99 & - \\
        \cmidrule(r){1-1}\cmidrule(r){2-4}\cmidrule(r){5-7}
        AVG  & 30.67 & 2.15 & 5.00 & 17.10 & 2.76 & - \\
        \bottomrule
        \multicolumn{7}{l}{$^*$FURIA builds different fuzzy sets for each rule.}\\
    \end{tabular}%
    \label{tab:complexity-small-data}%
\end{table}%

\vspace{5pt}
\subsubsection{Big vs Small Data algorithms on Small Data}\label{sssec:analysis-big-small-data}
\hfill\\
\indent In order to compare the classification performance of Big and Small Data algorithms, we carried out some non-parametric tests as recommended in the specialized literature~\cite{Garcia2008, Garcia2010}. More specifically, we used the Friedman's test~\cite{Friedman1940} to check whether there exist statistical differences among a group of methods and the Shaffer's \textit{post-hoc} test~\cite{Shaffer1986} to find the concrete pairwise comparisons which produce differences.

According to the Friedman's test, FURIA was the most accurate algorithm with a $p$-value of 4.779E-5, obtaining the following rankings (lower is better): FURIA (2.0000), FARC-HD (2.1667), FMDT (2.9444), FBDT (3.8889), CFM-BD (4.0000). The post-hoc Shaffer's test (Table \ref{tab:shaffer}) revealed that both FARC-HD and FURIA outperformed all Big Data algorithms with a significance level below 0.01 in all cases except for FMDT. When it comes to model complexities (Tables \ref{tab:complexity-big-small-data} and \ref{tab:complexity-small-data}), only CFM-BD and FBDT showed competitive results. Although FMDT was not statistically outperformed by FARC-HD and FURIA in terms of accuracy, the models built by this algorithm are much more complex than those built by FARC-HD and FURIA. These findings suggest that Big Data algorithms might not be as accurate as state-of-the-art Small Data algorithms when working with low complexity models, offering worse accuracy-complexity tradeoffs. However, there are some encouraging results in favor of Big Data algorithms that should be highlighted. In the case of \textit{Adult}, \textit{Pageblocks}, \textit{Shuttle}, and \textit{Wisconsin} datasets, they are not only able to achieve state-of-the-art accuracy rates, but also to build more compact models in the case of CFM-BD. This might be an indicator that we are on the right track towards a unified algorithm for Big and Small Data classification tasks. 
\begin{table}[!htbp]
    \centering
    \caption{adjusted $p$-values computed by shaffer's post-hoc test (underlined if $p$-value $\le0.01$)}
    \renewcommand\tabcolsep{3pt}
    \renewcommand{\arraystretch}{1.3}
    \begin{tabular}{@{}lccccc@{}}
    	\toprule
    	& CFM-BD & FBDT & FMDT & FARC-HD & FURIA\\
    	\midrule
    	CFM-BD vs. & - & 1.5037 & 0.2712 & \underline{0.0030} &\underline{0.0015}\\
        FBDT vs. & - & - & 0.2926 & \underline{0.0065} & \underline{0.0020}\\
        FMDT vs. & - & - & - & 0.4200 & 0.2926\\
        FARC-HD vs. & - & - & - & - & 1.5037\\
        FURIA vs. & - & - & - & - & -\\
        \bottomrule
    \end{tabular}
    \label{tab:shaffer}
    \vspace{-10pt}
\end{table}

\section{Discussion}\label{sec:discussion}
In the last few years, researchers have tried to design scalable fuzzy classifiers that reduce memory and computational requirements on Big Data classification tasks. However, these algorithms often rely on locally optimal subsolutions that might introduce approximation errors or make too strong assumptions about the training data. In this work, we carried out an empirical study to assess the extent to which the best-performing fuzzy classifiers for Big Data are sacrificing performance (both accuracy and model complexity) in favor of scalability. To this end, we compared these classifiers with state-of-the-art algorithms designed for standard classification tasks (Small Data). Since the latter were generally designed for maximizing performance without considering scalability issues, the results of this study provide some intuition around the tradeoff between performance and scalability achieved by Big Data solutions.

While Big Data algorithms are able to build compact fuzzy models with a similar complexity to that encountered in Small Data models, Big Data classifiers seem to be less accurate. These findings suggest we still need different approaches for Big and Small Data to achieve state-of-the-art classification accuracy. However, great progress has been made since the first distributed solutions based on the Chi et al. algorithm were proposed. Recent methods are gradually catching up with state-of-the art Small Data classifiers and show encouraging results. Of course, the experimental study carried out in this work has some limitations that should be considered. Distributed algorithms based on multiple locally optimal subsolutions (Chi-FRBCS-BigData and Chi-Spark-RS) usually run an independent optimization process on each data partition. As the number of partitions increases, the approximation error tends to increase and the quality of the global solution might drop. This performance loss could not be measured in our study and should be considered in future work.

\bibliographystyle{IEEEtran}

\end{document}